%% file: main.tex
\documentclass[11pt,a4paper]{amsart}

\title{Inference for max-linear Bayesian networks with noise}
\author{Mark Adams}
\address{Mark Adams, Naval Postgraduate School, Monterey}
\email{mark.p.adams@nps.edu}

\author{Kamillo Ferry}
\address{Kamillo Ferry, Technische Universität Berlin, Germany}
\email{ferry@math.tu-berlin.de}

\author{Ruriko Yoshida}
\address{Ruriko Yoshida, Naval Postgraduate School, Monterey}
\email{ryoshida@nps.edu}

\date{}
\keywords{max-linear Bayesian network, Gaussian mixtures, parameter inference}
\subjclass[2020]{14T90,62A09,62H30,90C20,90C90}

\usepackage[utf8]{inputenc}
\usepackage{amsmath,amssymb,amsthm,amsfonts,mathtools,color,url}
\usepackage{graphicx}

\usepackage{macros}
\usepackage{float}
\usepackage[headings]{fullpage}
\usepackage{booktabs}

\usepackage{tikz}
\usetikzlibrary{arrows,decorations.pathreplacing,shapes}
\usetikzlibrary{positioning,fit}
\usetikzlibrary{graphs,graphs.standard}
\usetikzlibrary{cd}

\tikzstyle{treenode} = [circle,minimum size=1cm,draw]
\usetikzlibrary{shapes}
\usetikzlibrary{arrows.meta}
\usetikzlibrary{patterns}
\usetikzlibrary{chains}
\usetikzlibrary{fit}
\usetikzlibrary{calc}

\usepackage{caption}
\usepackage{subcaption}

\usepackage{hyperref}

\usepackage[backend=biber,style=numeric,backref,backrefstyle=none]{biblatex}
\theoremstyle{plain}
\newtheorem{theorem}{Theorem}[section]

\newtheorem{lemma}[theorem]{Lemma}
\newtheorem{corollary}[theorem]{Corollary}

\theoremstyle{definition}
\newtheorem{definition}[theorem]{Definition}
\newtheorem{example}[theorem]{Example}

\theoremstyle{remark}
\newtheorem{remark}[theorem]{Remark}

\newcommand{\R}{\mathbb R}

\newcommand{\ppp}{\mathbb P}
\newcommand{\ttt}{\mathbf{T}}

\usepackage{dsfont}

\begin{document}
\begin{abstract}
Max-Linear Bayesian Networks (MLBNs) provide a powerful framework for causal inference in extreme-value settings; 
we consider MLBNs with noise parameters with a given topology in terms of the max-plus algebra by taking its logarithm. 
Then, we show that an estimator of a parameter for each edge in a directed acyclic graph (DAG) is distributed normally. 
We end this paper with computational experiments with the expectation and maximization (EM) algorithm and quadratic optimization. 
\end{abstract}

\maketitle
\section{Introduction}
Identifying and quantifying causal relationships is an objective in scientific inquiry and applied decision making processes. This objective becomes especially critical in the analysis of extreme events, which, despite their low frequency, can lead to disproportionately severe consequences in terms of cost and impact. Gaining insight into the underlying causal mechanisms is essential for informing risk management strategies, and guiding the development of effective mitigation policies. Max-linear Bayesian networks (MLBNs) have emerged as a powerful framework for studying causal relationships in extreme-value settings, where interactions between variables follow a max-linear structure \cite{GK:2018}.  These max-linear models have found applications in risk analysis, finance, and environmental sciences, where extreme observations drive decision-making. Examples include flooding events \cite{Causality_Extreme,TBK:2024a}, \cite{engelke2019graphicalmodelsextremes}, \cite{ExRN}, weather and climate \cite{Human_contribution}, and application to financial data \cite{Est_BN_scale} and to the European stock market \cite{ExFinance}.

A max-linear Bayesian network is a statistical model that is described by a 
weighted directed acyclic graph (DAG) in the following way.
Let \(G = (V,E)\) be a DAG with weight matrix \(C = (c_{ij})\in\R^{n\times n}_{\geq 0}\).
Then, the MLBN on \(G\) for a random vector 
\(X = (X_1,\dots,X_n)\) is defined by the recursive structural equations
\begin{equation}\label{eq:recursive-eqns2}
    X_j = \bigvee_{i \in \text{pa}(j)} c_{ij} X_i \vee c_{jj} Z_j, \quad i = 1, \dots, n,
\end{equation} where \(\vee\) denotes taking the maximum, \(\mathrm{pa}(j)\) denotes the \emph{parents} of vertex \(j\),
and $Z_1,\dots,Z_n$ are independent non-negative random variables called \emph{innovations}.
\begin{figure}[htb]
    \centering
    \input{figures/4node.tikz}
    \caption{A DAG consisting of four vertices. Each vertex $i$ in the network represents a random variable $X_i$ in the joint distribution of a random vector $X=(X_1, X_2, X_3, X_4)$.}
    \label{fig:4node}
\end{figure}
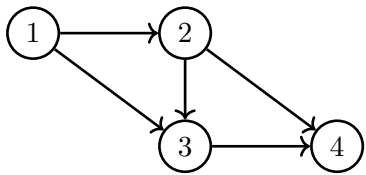

\begin{example}
    Figure~\ref{fig:4node}  represents a DAG on four vertices for a MLBN with a random variable $X=(X_1, X_2, X_3, X_4)$, a vector of innovations $Z = (Z_1, Z_2, Z_3, Z_4)$ and a weight matrix \({C\in\R^{4\times 4}_{\geq 0}}\). The structural equations \eqref{eq:recursive-eqns2} for the model are given by:
    \[
    \begin{aligned}
        X_1 &= Z_1,\\
        X_2 &= c_{12}X_1 \vee Z_2,\\
        X_3 &= c_{13}X_1 \vee c_{23}X_2 \vee Z_3,\\
        X_4 &= c_{24}X_2 \vee c_{34}X_3 \vee Z_4,
    \end{aligned}
    \quad\text{where}\quad
    C = \begin{pmatrix}
         1&c_{12}&c_{13}& 0  \\
         0&1&c_{23}&c_{24}\\
         0 & 0 & 1 & c_{34}\\
         0 & 0 & 0 & 1
    \end{pmatrix}.\]
    Explicitly writing down the solution given by the matrix $C$ allows us to express
    the random vector \({X = (X_1,X_2,X_3,X_4)}\) by 
    \begin{equation}\label{eq:example1}
        \begin{aligned}
            X_1 &= Z_1,\\
            X_2 &= c_{12}Z_1 \vee Z_2,\\
            X_3 &= (c_{13} \vee c_{12}c_{23}) Z_1 \vee c_{23}Z_2 \vee Z_3,\\
            X_4 &= (c_{12}c_{24} \vee c_{13}c_{14} \vee c_{12}c_{23}c_{34}) Z_1
                \vee (c_{24} \vee c_{23}c_{34})Z_2 \vee c_{34} Z_3 \vee Z_4.
        \end{aligned}
    \end{equation}
    In other words, equation~\eqref{eq:example1} describes a matrix-vector equation
    \(X = Z\cdot C^*\) where
    \[
        C^* = \begin{pmatrix}
             1&c_{12}&c_{13} \vee c_{12}c_{23} & c_{12}c_{24} \vee c_{13}c_{14} \vee c_{12}c_{23}c_{34} \\
             0&1&c_{23}&c_{24} \vee c_{23}c_{34}\\
             0 & 0 & 1 & c_{34}\\
             0 & 0 & 0 & 1
        \end{pmatrix}.
    \]
\end{example}

A central challenge in the analysis and application of MLBNs lies in the estimation of parameters, matrix $C$, 
due to the model's tropical structure. 
These parameters act as multiplicative weights along the directed edges of the network, while the vertices assume values through max-linear operations. 
As a result of max-linear operations, standard likelihood based estimation techniques are not directly applicable \cite{KL:2019}. 
In particular, \citeauthor{GKL:2019}~\cite{GKL:2019} are able to identify possible edge weights 
with a sufficient number of samples, and without noise in the model. 
\citeauthor{Buck2020RecursiveMM}~\cite{Buck2020RecursiveMM} derive estimators under the assumption 
of one sided noise, specifically when $E \geq 1$. 
Inspired by the Latent Tree problem \cite{tran2022tropicalgeometrycausalinference}, and sensor collection error, 
we develop a statistical framework for parameter estimation under more relaxed noise constraints.

\subsection{Problem Statement}
We develop a statistical estimation framework for the parameter matrix of a MLBN 
in the presence of multiplicative noise, assuming that the underlying DAG structure is known. 
For this, we introduce a modification to the standard max-linear recursive equations~\eqref{eq:recursive-eqns}
by incorporating a strictly positive random variable $E_j > 0$ with continuous atom-free distribution into each structural equation. The resulting model takes the form
\begin{equation}\label{eq:noisy-recursive-eqns}
    X_j = \bigvee_{i \in \text{pa}(j)} (c_{ij} X_i \vee c_{jj} Z_j)E_j, \quad i = 1, \dots, n,
\end{equation} for a matrix $C = (c_{ij}) \in \mathbb{R}_{\geq 0}^{n \times n}$, and $E_j$ represents multiplicative noise, capturing variability that distorts the observed values of $X_j$.

Our goal is to develop statistically sound and computationally efficient inference procedures that leverage the algebraic structure of the max-times semiring and the sparsity inherent in the DAG, enabling accurate estimation of $C$ from observed data subject to noise.

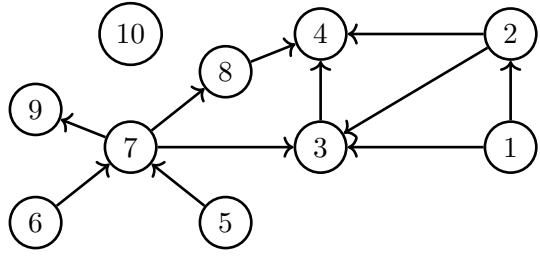
\begin{figure}[tb]
    \centering
    \input{figures/10node.tikz}
    \caption{This curated network provides the basis for our analysis throughout the paper. The unique combinations of directed edges creates triangles, diamonds, and max-weighted paths that create challenges in parameter estimation.}
    \label{fig:10node}
\end{figure}

\begin{figure}[b]
    \centering
    \begin{subfigure}{0.3\textwidth}
        \centering
        \begin{tikzpicture}[line width=1pt,x=1.5cm,y=1cm]
            \node at (0,-2.75) {}; 
            \node[draw, circle] (1) at (0,1) {$1$};
            \node[draw, circle] (2) at (-1,-.5) {$2$};
            \node[draw, circle] (3) at (1,-.5) {$3$};
            \node[draw, circle] (4) at (0,-2) {$4$};
            
            \draw[->] (1) --node[anchor=south east]{$c_{12}$} (2);
            \draw[->] (1) --node[anchor=south west]{$c_{13}$} (3);
            \draw[->] (2) --node[anchor=south]{$c_{23}$} (3);
            \draw[->] (2) --node[anchor=north east]{$c_{24}$} (4);
            \draw[->] (3) --node[anchor=north west]{$c_{34}$} (4);
        \end{tikzpicture}
        \caption{Triangular Structure}
        \label{fig:graph2}
    \end{subfigure}
    \hfill
    \begin{subfigure}{0.3\textwidth}
        \centering
        \begin{tikzpicture}[line width=1pt,x=1.5cm,y=1cm]
            \node at (0,-2.75) {}; 
            \node[draw, circle, line width=1pt] (1) at (0,1) {$7$};
            \node[draw, circle] (2) at (-1,-.5) {$3$};
            \node[draw, circle] (3) at (1,-.5) {$8$};
            \node[draw, circle] (4) at (0,-2) {$4$};
            
           \draw[->] (1) --node[anchor=south east]{$c_{37}$} (2);
            \draw[->] (1) --node[anchor=south west]{$c_{78}$} (3);
            \draw[->] (2) --node[anchor=north east]{$c_{34}$} (4);
            \draw[->] (3) --node[anchor=north west]{$c_{84}$} (4);
        \end{tikzpicture}
        \caption{Diamond Structure}
        \label{fig:graph3}
    \end{subfigure}
    \hfill
    \begin{subfigure}{0.3\textwidth}
        \centering
        \begin{tikzpicture}[line width=1pt,x=1.5cm,y=1cm]
            \node at (0,-2.75) {}; 
            \node[draw, circle, line width=1pt] (1) at (-1,1) {$5$};
            \node[draw, circle] (2) at (1,1) {$6$};
            \node[draw, circle] (3) at (0,-.5) {$7$};
            \node[draw, circle] (4) at (0,-2) {$9$};
            
            \draw[->] (1) --node[anchor=north east]{$c_{57}$} (3);
            \draw[->] (2) --node[anchor=north west]{$c_{67}$} (3);
            \draw[->] (3) --node[anchor=west]{$c_{79}$} (4);
        \end{tikzpicture}
        \caption{Y-Structure}
        \label{fig:graph4}
    \end{subfigure}
    \caption{Graph structures our network integrates to examine parameter estimation in combination and isolation.}
    \label{fig:GraphStructures}
\end{figure}

\subsection{Motivation}
In 2021, Gissibl et al.~\cite{GKL:2019} established the identifiability of max-linear recursive equations in the noise-free setting by exploiting the structure of consistently occurring observations. To the best of our knowledge, the problem of parameter estimation in a  setting with noise, particularly multiplicative noise, has not been systematically addressed by literature. A key challenge in this setting is the disentanglement of the multiplicative noise component from the underlying distribution, especially when noise obscured the max-linear dependencies. Developing accurate and robust estimators for the edge weights of the MLBN is crucial, as it enables more reliable inference of causal pathways, and improves our ability to reason about the causal impacts of extreme events in complex systems. 

Understanding the causal relations between variables enables informed decisions about the effects of interventions, which is essential in informed policy design.
In the context of extreme events and MLBNs, the model's structure implies that system behavior is governed by dominant risk pathways. Consequently, effective policies should focus on mitigating the most influential risk factors.

To investigate the behavior of parameter estimation under structural assumptions, we consider the network depicted in Figure~\ref{fig:10node}, which incorporates several substructures commonly encountered literature \cite{Amendola:2022} \cite{LEGSiHD}. These are highlighted in Figure~\ref{fig:GraphStructures}. Subgraph I exhibits a double triangle structure, which under certain parameter configuration render parts of subgraph I's structure unidentifiable \cite{GKL:2019}. Specifically in any triangle the parameter $c_{13}$ is unrecoverable when $c_{13}<c_{12}c_{23}$ \cite{GKL:2019}. In a double triangle, the max weighted path can mask parameters. Subgraph II corresponds to a diamond structure, where max flow will dominate, but long tailed distributions allow for full recovery, and Subgraph III presents a Y-structure that facilitates simple yet detailed analysis. 

Together these subgraphs provide the opportunity to conduct parameter estimation under a multitude of conditions. Additionally, we study each of these motifs in isolation to examine edge cases and structural challenges that may arise during parameter estimation. 
Furthermore, we introduce an independent vertex, labeled node $10$, to facilitate the investigation of identifiability conditions associated with marginal independence, and aspect we intend to explore in future work.

\subsection{Contributions of this paper}

This work contributes to the statistical foundations of MLBN by addressing parameter estimation in structured graphical models under noise. We demonstrate how logarithmic transformations can be leveraged to estimate parameters in the presence of noise and we provide a theoretical justification for using Gaussian Mixture Models (GMMs) as a statistical estimation tool within a MLBN framework. 

In parallel, we propose a quadratic optimization problem for estimating parameters for MLBN as an alternative. For an expert in tropical geometry, note that under our quadratic optimization problem, a feasible region forms a {\em polytrope}, which is a classical convex hull and a tropical convex hull (see \cite{JK:2010} for details on polytropes).  Our problem becomes estimating the \emph{Kleene star} $C^*$ from the observations.

We conduct a comparative analysis a GMM-based estimator and a tropical hyperplane-based approach highlighting the respective strengths and limitations of each method.
In particular, we empirically investigate tail dependency in the context of structural inactivation, examining its consequences for both approaches. 
Our findings indicate that the GMM-based estimator exhibits inconsistency, becoming unstable under tail dependency.
Furthermore, under a fixed noise experiment, we demonstrate the performance of the GMM-based estimator
with respect to the number of required observations, and demonstrate how tropical methods remain effective.

\subsection{Organization of this paper}
Our paper is organized as follows. In Section~\ref{sec:preliminaries}, we provide an overview 
of relevant key concepts, including graph terminology, Bayesian networks, tropical geometry, and max-linear models.
Section~\ref{sec:param-estimation} establishes the justification for using GMMs in parameter estimation. 
We formally define the estimation problem, introduce log-space representations, and lay the foundation for 
why GMMs are a suitable choice for inference in max-linear models. 
We discuss the conditions under which GMM-based estimation fails, and how hyperplane-based estimation may be a 
better option. 
In Section~\ref{sec:experiments}, we present analytical results, and provide a comparison of GMM and tropical hyperplane methods, 
highlighting differences in efficiency and interpretability. 
Section~\ref{sec:conclusion} concludes the paper and summarizes our key finding discussing their implications and providing 
the basis for future research directions.

\section{Preliminaries}\label{sec:preliminaries}

\subsection{Graph Terminology}
First, we fix the basic terminology regarding graph theory we are going to use.
In the following, we study simple directed acyclic graphs (DAG), $G = (V, E)$, defined by a finite sets 
of vertices ${V = [n] \coloneqq \{1, \dots, n\}}$ and edges $E\subset V\times V$. 
In the terminology of \citeauthor{Lauritzen:2004}~\cite{Lauritzen:2004} we consider pure graphs only consisting of directed edges.

An edge $e\in E$ is defined by its \emph{source} \(i\) and \emph{target} \(j\).
This way, we keep to the same graph notation as \citeauthor{Amendola:2022}~\cite{Amendola:2022} 
except $i$ becomes the parent and $j$ becomes the child given an edge \(i\to j\). 
A path \(i\rightsquigarrow j\) in \(G\) is defined as a sequence of distinct nodes $d_0, d_1, \ldots , d_{\ell}$
\( (i = d_0, d_1, \dots, d_{\ell} = j) \) such that \(d_k\to d_{k+1}\) is an edge in \(G\) for all \(0\leq k < \ell\).

The set of \emph{parents} of $j$ is denoted $\parents(j)$ and the set of \emph{children} of $i$ is $\children(i)$. 
These relationships can further be categorized into ancestors and descendants. 
Here $j$ is a \emph{descendant} of $i$ and $i$ is an \emph{ancestor} of $j$ if there exists a path from $i$ to $j$, denoted by $i\rightsquigarrow j$. We denote the set of ancestors by \(\ancestors(i)\)
and define the set of extended ancestors as $ \overline{\ancestors}(i) = \ancestors(i) \cup \{ i \}$. 

A weighted directed graph is a simple directed graph together with a weight matrix 
\(C\in\R_{\geq 0}^{n\times n}\) such that \(c_{ii} = 1\) and \(c_{ij} > 0\) whenever \(i\to j \in E\).

\subsection{Probability primer}
Let \((\Omega,\mathcal{A},\ppp)\) be a probability space.
The distribution of a random variable \(X\colon\Omega\to\R\) is the image measure of \(\ppp\) under \(X\), i.\,e.\ 
the probability measure \(\ppp_X\) on \(\R\) with the Borel \(\sigma\)-algebra defined by \[
    \ppp_X(B) \coloneqq \ppp(X^{-1}(B))
\] where \(B\subset\R\) is a Borel set. The smallest closed set \(B\subset\R\) such that \(\ppp_X(B) = 1\)
is called the \emph{support} of \(X\).

We say that a probability measure \(\ppp\) on a measurable space \((\Omega,\mathcal{A})\) has 
\emph{density} \(f\) if there exists another measure \(\lambda\) on \(\Omega\) such that \[
    \ppp(A) = \int \mathbf{1}_A f \,\mathrm{d}\lambda.
\] If \(f\) is the density of \(\ppp_X\), we say that the random variable \(X\) has density \(f\).

A Borel set \(B\subset\R\) is called an \emph{atom} for the probability measure \(\ppp\) if 
\(\ppp(B) > 0\) and the measure of every Borel set \(A\subset B\) is either \(0\) or \(\ppp(B)\).
We say that \(x\in\R\) is an atom for the random variable \(X\colon\Omega\to\R\) if \(\{x\}\) is an atom of 
\(\ppp_X\).

\subsection{Bayesian Networks}
We give a overview of Bayesian networks mostly following \citeauthor{Sullivant:2018}~\cite{Sullivant:2018}.
Fix a DAG \(G = (V,E)\) with \(V = [n]\). We say that a distribution for 
a random vector \({X = (X_1,\dots,X_n)}\) is \emph{Markov relative to \(G\)} if the density of \(X\) 
factors into the conditional densities \begin{equation}\label{eq:markov-relative-density}
    f(x) = \prod_{j\in V} f_j(x_j \mid x_{\parents(j)})
\end{equation} where \(f_j(\cdot \mid x_{\parents(j)})\) is the marginal density of \(X_j\) conditioned 
on the parents \(\parents(j)\).
The \emph{Bayesian network} on \(G\) is the statistical model consisting of all probability densities Markov relative to \(G\).
Another important characterization of Bayesian networks can be stated in terms of \emph{Markov properties}.
We say the \emph{directed local Markov property} associated to \(G\) consists of all conditional independence statements of the form
\begin{equation}\label{eq:local-Markov-property}
    X_i \ci X_{\nondescendants(i)\setminus\parents(i)} \mid X_{\parents(i)}
\end{equation} for \(i\in V\). Then, for a density to be Markov relative to \(G\) is equivalent to satisfying the 
directed local Markov property associated to \(G\), i.\,e.\ satisfying all the CI statements in \eqref{eq:local-Markov-property} \cite[Thm.\ 13.2.10]{Sullivant:2018}.

\subsection{Tropical semirings and polytropes}
Max-linear Bayesian networks are inherently tropical objects.
For this, we introduce the necessary preliminaries from tropical geometry to make our setting precise.

There are two tropical semirings that are relevant for us, the \emph{max-times semiring} \(\R_{\geq 0}\)
equipped with operations 
\[
a \vee b:= \max(a,b),  \quad  a \cdot b:= ab  \quad \text{for } a,b \in \R_{\geq 0} \coloneqq [0, \infty).
\] and the \emph{max-plus semiring} $(\ttt, \oplus, \odot)$ where
\[
a \oplus b:= \max(a,b),  \quad  a \odot b:= a + b  \quad \text{for } a,b \in \ttt\coloneqq\R\cup\{-\infty\}.
\] Multiplication of matrices over these semirings is carried out analogously to the classical case using 
the corresponding addition and multiplication of the semiring.

Denote by \(\ttt\mathbb{A}^{d-1}\) the \emph{tropical affine space} which is the 
set of points \(x\in\R^d\) identified via the equivalence relation \[
    (x_1,\dots,x_n)\sim \lambda\odot(x_1,\dots,x_d)=(x_1+\lambda,\dots,x_d+\lambda)
\] for all \(\lambda\in\R\). This is also sometimes called \emph{tropical projective torus} in the literature. 

We can also interpret \(\ttt\mathbb{A}^{d-1}\) as the quotient \(\R^d/\R\mathbf{1}\)
of Euclidean space by the subspace spanned by the all-ones vector \(\mathbf{1} = (1,\dots,1)\).
By fixing a coordinate to be \(0\), that is \[
    (x_1,x_2,\dots,x_d) \sim (0, x_2 - x_1, \dots x_d - x_1),
\] tropical affine space \(\ttt\mathbb{A}^{d-1}\) is homeomorphic to \(d-1\)-dimensional Euclidean space \(\R^{d-1}\).

\begin{remark}\label{rem:max-plus-vs-max-times}
    The semirings \(\R_{\geq 0}\) and \(\ttt\) are isomorphic by taking the logarithm resp.\ exponentiation.
    While max-linear Bayesian networks will be defined over the max-times semiring, 
    only geometry over the max-plus semiring allows for the necessary comparison to Euclidean geometry.
    This way, we also consider weighted directed graphs with weight matrix \(C\in\ttt^{n\times n}\) 
    over the max-plus semiring such that \(c_{ii} = 0\) and \(c_{ij} > -\infty\) whenever \(i\to j\in E\).
\end{remark}

Let \(C = (c_{ij})\in\ttt^{n\times d}\) be a tropical matrix such that tropical sums of rows and columns are finite.
Such a matrix is called \emph{\(\R\)-astic} following \citeauthor{Butkovic:2010}~\cite{Butkovic:2010}. 

\begin{definition}
A \emph{tropical polytope} is defined as the row space of a tropical \(\R\)-astic matrix 
\(C\in\ttt^{n\times d}\), so \[
    \tconv(C) = \left\{\,
        x\odot C \mid x\in\R^n
    \,\right\} \subseteq \ttt\mathbb{A}^{d-1}.
\]
\end{definition}

In the special case where \(C\) is a \(\R\)-astic square matrix is idempotent and 
has zero-diagonal, we say that \(C\) is a \emph{Kleene star}. 
In particular, for any \(\R\)-astic square matrix \(C\in\ttt^{n\times n}\)
the expression \begin{equation}\label{eq:kleene-star}
     C^* = I_n \oplus C \oplus C^{\odot 2} \oplus \dots \oplus C^{\odot(n-1)} \oplus \dots
\end{equation} is the \emph{Klenee star} associated to \(C\) if and only if \(C\) is the weight matrix of
a directed graph without positive cycles.

\citeauthor{JK:2010}~\cite{JK:2010} showed that the tropical polytope of a Kleene star is classically convex 
under above identification of \(\ttt\mathbb{A}^{d-1}\) with \(\R^{d-1}\). 
They also coined the term \emph{polytrope} for this type of tropical polytope. In particular, 
a polytrope is a tropical simplex, thus \(d = n\).

Since any polytrope \(P = \tconv(C^*)\) is also a classical polytope, it has a classical facet description, 
which is \begin{equation}\label{eq:wdp}
    \mathrm{Q}(C) = \left\{\,
        x\in\R
        \mid
        x_j - x_i \geq c_{ij}
    \,\right\}.
\end{equation} It turns out that 
\[
    \mathrm{Q}(C) = \mathrm{Q}(C^*) = \tconv(C^*) \subseteq \tconv(C)
\] where the last containment is strict unless \(C = C^*\) \cite{De-La-Puente:2013}.

\begin{remark}\label{rem:weighted-transitive-reduction}
\citeauthor{AF:2024}~\cite{AF:2024} characterized for DAGs \(G\) the perturbations of weight matrices \(C\) preserving 
the associated Kleene star \(C^*\). This happens in terms of the optimal transport problem on \(G\). That is,
a hyperplane \(\{\,x_j - x_i = c_{ij}\,\}\) defines a facet of the polytrope \(\mathrm{Q}(C)\) if and only if the edge \(i\to j\)
is the unique optimal path connecting \(i\) to \(j\) \cite[Thm. 4.9]{AF:2024}.
This means that edges in \(G\) might become irrelevant depending on the weights \(C\).
\end{remark}

\subsection{Recursive structural equations and max-linear Bayesian networks}
We turn to max-linear Bayesian networks which are a \emph{recursive structural equation model} over the 
max-times semiring introduced by \citeauthor{GK:2018}~\cite{GK:2018}. 

\begin{definition}
For a weighted DAG \(G = (V,E)\) with weight matrix \(C\in\R^{n\times n}_{\geq 0}\) define the 
\emph{max-linear Bayesian network} (MLBN) \(X = (X_1,\dots, X_n)\) by the equations
\begin{equation}\label{eq:recursive-eqns}
    X_j = \bigvee_{i \in \text{pa}(j)} c_{ij} X_i \vee c_{jj} Z_j, \quad i = 1, \dots, n,
\end{equation} where $Z=(Z_1,\dots,Z_n)$ are assumed to be independent random variables, each with support 
\(\R_{>0} = (0,\infty)\) and atomfree distributions.
\end{definition}
We may write above recursive linear system of equations compactly as the max-times matrix-vector product \[
    X = X\cdot C \vee Z.
\] By repeated substitution, this recursive equation system admits the solution 
\(X = Z\cdot C^*\) where \(C^*\) is the Kleene star over the max-times semiring.
By assumption, \(C\) is the weight matrix of a directed acyclic graph making \(C^*\) well-defined.

After applying a logarithmic transformation, the set of possible observations for \(\log{X}\) forms a polytrope in 
\(\ttt\mathbb{A}^{n-1}\). For this reason, we discuss the properties of \(\log{X}\) from now on. 

If \(\omega\coloneqq \log{C^*}\) denotes the logarithmic Kleene star of the weight matrix for the MLBN \(X\), 
it follows from \eqref{eq:wdp} that the observations of the difference \[Y_{ij} \coloneqq \log{X_j} - \log{X_i}\] 
will be bounded from below.

In general, describing the distribution of a MLBN is infeasible, 
since the random variables \(\log{X_j}\) arise as weighted maxima of the random variables \(\log{Z_i}\).
\citeauthor{KL:2019} characterized the atoms of the random variable \(Y_{ij}\), that is the values 
\(x\in\ttt\mathbb{A}^{n-1}\) for \(X\) ocurring with positive probability.

\begin{lemma}[{\cite[Lemma 3.4]{GKL:2019}}]\label{lem:atoms_of_differences}
    Let \(i\neq j\in V(G)\) be distinct nodes of the underlying graph \(G\). Then, the random variable \(Y_{ij}\)
    has an atom at \(\omega_{kj} - \omega_{ki}\) for every common ancestor \(k\) and these are the only atoms.
    In particular, if \(i\) is an ancestor of \(j\), then there is an atom at \(\omega_{ij}\).
\end{lemma}
As a consequence of Lemma~\ref{lem:atoms_of_differences}, for a sample \(Y_{ij}^1,\dots, Y_{ij}^N\) 
of differences \(Y_{ij}\) without noise, the estimator \begin{equation}\label{eq:KL1}
  \hat\omega_{ij} = \min_{\nu=1}^N\left(Y_{ij}^\nu\right)
\end{equation} will be exactly equal to the true parameter with high probability \cite[Example 6]{KL:2019}.
In particular, standard likelihood functions are not well-defined in this setting because 
there are no known density functions for distributions involving the maximum operator over 
generalized extreme value distributions. This means that the estimator in \eqref{eq:KL1} is a 
\emph{generalized MLE}, for details see \cite{GMLE_Kiefer_W}.

\begin{remark}\label{def:structural-inactivation}
    The phenomenon discussed in Remark \ref{rem:weighted-transitive-reduction} applies to max-linear Bayesian 
    networks, particularly when an edge is either removed or rendered functionally insignificant within the network. 
    \emph{Structural inactivation} of the edge occurs when $\mathbb{P}_{c_{ij}}(X_j = X_ic_{ij}) =0$.
    
    Due to the statistical nature, structural inactivation can occur due to substantial reduction in the weight of parameter $c_{ij}$. 
    In our setting, we define an edge to be approaching structurally inactivation when $\mathbb{P}_{c_{ij}}(X_j = X_ic_{ij}) < 0.05$. The threshold of 0.05 is adopted due to its conventional use in capturing tail dependence, as well as its empirical relevance as demonstrated by our observations in Table~\ref{tab:observations}. 
    When noise is present in the model, a large sample is required to estimate $c_{ij}$ due to the minimal probabilistic likelihood of an observation occurring along that edge.

    The triangular structures depicted in Figure \ref{fig:graph2} impedes the detection of structural inactivity in edge $c_{13}$. In particular, $c_{13}$ inherits flow characteristics from the path  $c_{12}c_{23}$ once its capacity is exceeded, conflating direct and indirect transmission effects and obscuring the contribution of $c_{13}$.
\end{remark}

\section{Parameter estimation under uncertainty}\label{sec:param-estimation}
In this section, we study the problem of estimating the max-linear coefficients for a fixed DAG under the assumption of Gaussian noise. 
We leverage established estimation techniques and make use of the graph structure from Figure~\ref{fig:10node} for analysis, which will serve as the foundation for this and the following section. 

Assume that we are given observations of a MLBN \(X = (X_1,\dots, X_n)\) 
with the presence of noise \(E_j\) log-normally distributed. 
This means in particular that \(\varepsilon_j\coloneqq\log{E_j}\sim N(0, \sigma_j^2)\)
with $\sigma_j > 0$, which is a continuous atom-free distribution.
Following Remark \ref{rem:max-plus-vs-max-times}, we decide to work over the max-plus semiring.
Thus, the problem we study is to estimate the parameters $\omega_{ij}$ 
given data with noise that satisfies the equations
\begin{equation}\label{eqn:log-MLBN}
 \log{X_j}\odot\varepsilon_j = 
    \left(\bigoplus_{i \in \text{pa}(j)}
         \log{c_{ij}} \odot  \log{Z_i}
        \oplus  \log{c_{jj}} \odot  \log{Z_{j}}  
    \right) \odot \varepsilon_j .
\end{equation}

\subsection{Gaussian Mixture Models}
Each random variable \(X_j\) arises as the maximum over several weighted random variables. 
We can see this as one specific observation for \(\log{X_j}\) being selected at random from the 
expressions \(\omega_{ij}+ \log(Z_i)\) for each path from \(i\) to \(j\) in the underlying graph.
In this section, we elaborate how in the setting with noise, above observation leads to the application of Gaussian mixture models.

\begin{definition}
A \emph{mixture} is a random variable \(X\) with density \(f\) given by the convex combination
of probability densities \(f_k\), that is \[
    f(x) = \sum_{k=1}^{K} \pi_k f_{k}(x)
\] where \( K \) is the number of \emph{mixture components}
and \( \pi_k \geq 0 \) are the \emph{mixing weights} satisfying~\( \sum_{k=1}^{K} \pi_k = 1 \).
If \(D_k\) are probability distributions with density \(f_k\), we may denote \(X\) being a mixture 
by \(X\sim\sum_{k=1}^K \pi_KD_k\).

If for each \({1\leq k\leq K}\), \(f_k\) is the density of a Gaussian distribution with mean \(\mu_k\) and variance \(\sigma_k^2\) we say that \(X\) is a \emph{Gaussian mixture}.
In this case, we write \(X\sim\sum_{k=1}^K \pi_k N(\mu_k,\sigma^2_k)\).
\end{definition}

Under conditions with noise, the discrete atoms of \( Y_{ij} \) become normally distributed.
This suggests that we may see the differences \(Y_{ij}+(\varepsilon_j - \varepsilon_i)\)
as distorted Gaussian mixtures in the following way.

\begin{theorem}\label{lem:distribution-of-coordinate-differences}
    Assume that $\varepsilon_j\sim N(0, \sigma_j^2)$ with $\sigma_j > 0$ for \(j\in V(G)\).
    Then, there exists a distribution \(D\) and real numbers \(0 \leq \pi_k,\pi \leq 1\) for every common
    ancestor \(k\) of \(i\) and \(j\) with \(\pi+\sum_k \pi_k = 1\) such that \(Y_{ij}\) 
    has as distribution the following finite mixture \[
        Y_{ij} + (\varepsilon_j - \varepsilon_i)
        \sim 
        \sum_{k\in\overline{\ancestors}(i)\cap\overline{\ancestors}(j)} 
            \pi_k N( \omega_{kj} - \omega_{ki}, \sigma_i^2 + \sigma_j^2)
        + \pi D.
    \]

    \begin{proof}
        It follows from Lemma \ref{lem:atoms_of_differences} that the distribution of \(Y_{ij}\) decomposes
        into \[
            Y_{ij} \sim \sum_{k\in\overline{\ancestors}(i)\cap\overline{\ancestors}(j)} 
                \pi_k \delta_{\omega_{kj} - \omega_{ki}}
                + \pi D'
        \] where \(\delta_c\) denotes the \emph{Dirac distribution} at \(c\in\R\) and 
        \(D'\) is a distribution that agrees with the distribution of \(Y_{ij}\) outside of the atoms.
        Note that each random variable \(X_j\) may be expressed as \[
            \log{X_j} = \bigoplus_{\ancestors(j)}\left(
                    \log{Z_i} \odot \omega_{ij}
                \right) \oplus\log{Z_j}
        \] which means that every observation of \(X_j\) and \(X_i\) is realized by a specific term.
        Thus, the atom \(\omega_{kj} - \omega_{ki}\) of \(Y_{ij}\) is observed when \(Z_k\gg 0\) because \[
            Y_{ij} = (\omega_{kj} + Z_k) - (\omega_{ki} + Z_k) = \omega_{kj} - \omega_{ki}.
        \] Since the noise \(\varepsilon_j - \varepsilon_i\) is additive, \(Y_{ij}\vert_{Z_k\gg 0}\)
        follows a normal distribution \({N( \omega_{kj} - \omega_{ki}, \sigma_i^2 + \sigma_j^2)}\) centered around 
        \(\omega_{kj} - \omega_{ki}\). Since the atoms in the distribution of \(Y_{ij}\) correspond exactly to 
        the common ancestors \(k\) of \(i\) and \(j\), the statement follows.
    \end{proof}
\end{theorem}

As a consequence of Lemma~\ref{lem:atoms_of_differences} and Theorem~\ref{lem:distribution-of-coordinate-differences},
when approximating \(Y_{ij} + (\varepsilon_j - \varepsilon_i)\) by a Gaussian mixture,
that the leftmost mixture component corresponds to the value of \(\omega_{ij}\) we are interested in.
That is, \(\omega_{ij} = \min_k\{ \mu_k \}\) where \(\mu_k = \omega_{kj} - \omega_{ki}\) are the means
of the Gaussian mixture in Theorem~\ref{lem:distribution-of-coordinate-differences}.
This leads to the following estimator.

\begin{corollary}\label{cor:smallest-peak}
     Assume that $\varepsilon_j\sim N(0, \sigma_j^2)$ with $\sigma_j > 0$ for \(j\in V(G)\)
     and let \(X^1,\dots,X^N\) be an i.i.d.\ sample of the max-linear Bayesian network.
     If \(i\) is an ancestor of \(j\), then \[
     \hat{\omega}_{ij} = \min_\nu(Y_{ij}^\nu) + \varepsilon_j - \varepsilon_i \sim N(\omega_{ij}, \sigma_i^2 + \sigma_j^2).
     \] In particular, \(N(\omega_{ij}, \sigma_i^2 + \sigma_j^2)\) occurs as a component of the 
     mixture \(Y_{ij} + (\varepsilon_j - \varepsilon_i)\).

     \begin{proof}
        By Lemma \ref{lem:atoms_of_differences}, the distribution of \(Y_{ij}\) contains an atom at \(\omega_{ij}\)
        if \(i\) is an ancestor of \(j\). By \eqref{eq:wdp} this is in particular the minimum of the
        support of \(Y_{ij}\).
        It follows from Theorem \ref{lem:distribution-of-coordinate-differences} that under the presence of noise
        this atom is replaced by the component \(N(\omega_{ij}, \sigma_i^2 + \sigma_j^2)\).
     \end{proof}
\end{corollary}

\begin{example}\label{ex:GMM}
\begin{figure}[bt]
    \centering
    \begin{subfigure}{0.3\textwidth}
        \centering
        \begin{tikzpicture}[scale=1.6, line width=1pt,y=.75cm]
            \node[draw, circle, line width=1pt] (1) at (-1,0) {$1$};
            \node[draw, circle] (2) at (0,0) {$2$};
            \node[draw, circle] (3) at (1,0) {$3$};
            \node[draw, circle] (4) at (0,-2) {$4$};
            
            \draw[->] (1) --node[anchor=north east]{$t^3$} (4);
            \draw[->] (2) --node[anchor=west, pos=.25]{$t^{1.5}$} (4);
            \draw[->] (3) --node[anchor=north west]{$t^2$} (4);
        \end{tikzpicture}
        \caption{Graph structure}
        \label{fig:graph}
    \end{subfigure}
    \hfill
    \begin{subfigure}{0.33\textwidth}
        \centering
        \includegraphics[trim={0 0 0 2cm}, clip, width=\linewidth]{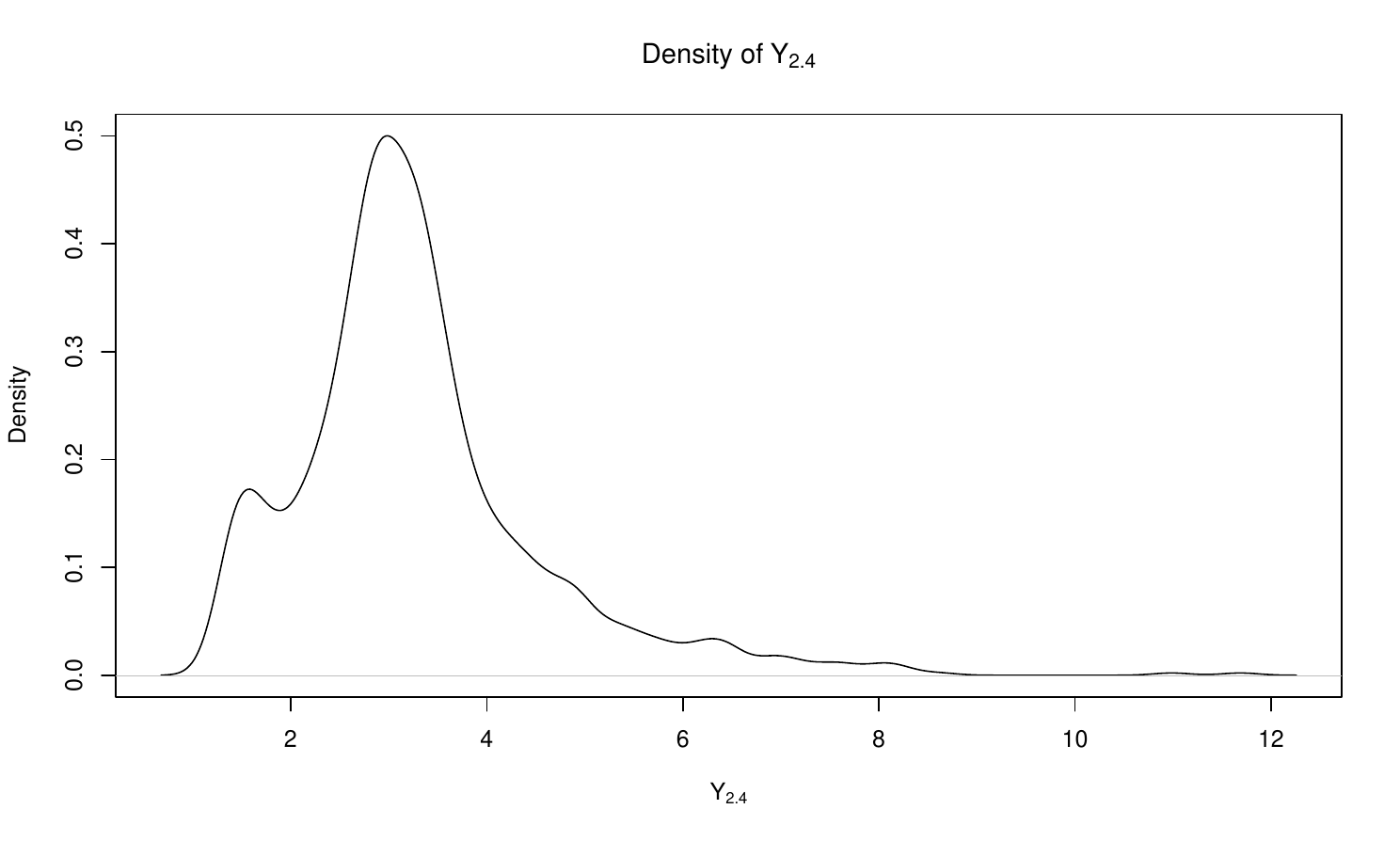}
        \caption{Density plot of $Y_{24}$}
        \label{fig:density}
    \end{subfigure}
    \hfill
    \begin{subfigure}{0.33\textwidth}
        \centering
        \includegraphics[trim={0 0 0 2cm}, clip, width=\linewidth]{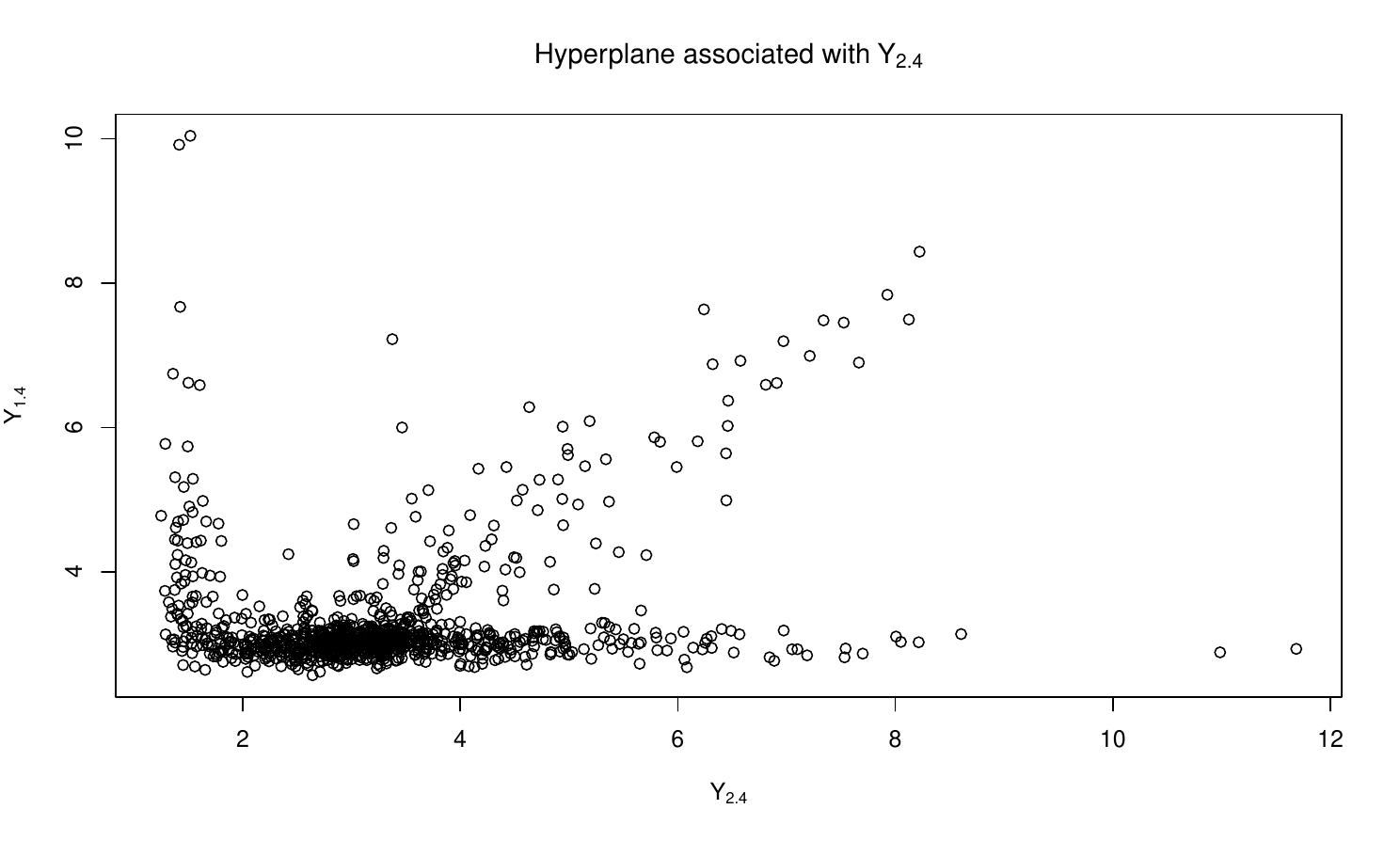} 
        \caption{Marginal plot of \(Y_{14}\) vs.\ \(Y_{24}\)}
        \label{fig:Polytrope}
    \end{subfigure}
    \caption{A max-linear Bayesian network on 4 nodes with $N = 2000$ along with the density plot of $Y_{24}$ and the marginal plot of \(Y_{14}\) vs.\ \(Y_{24}\). These visualizations provide insights into the effectiveness of our methodology, naming the application of GMM and the geometric of the associated polytrope.}
    \label{fig:three_figures}
\end{figure}
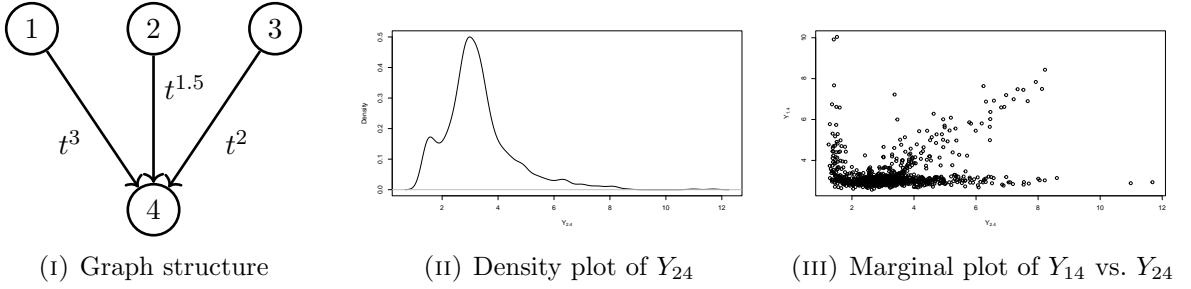

Figure \ref{fig:three_figures} shows an example of a random sample generated from the logarithms of the MLBN 
with Gaussian noise $N(0,0.1)$ for all \(j\in V(G)\). The logarithmic weights of the MLBN are given by \[
    \omega = \log{C} = \begin{pmatrix}
         0 & -\infty & -\infty & 3\\
         -\infty & 0 & -\infty & 1.5\\
         -\infty & -\infty & 0 & 2\\
         -\infty & -\infty & -\infty & 0
     \end{pmatrix}
\] Knowing the structure of the network, we expect in accordance with Lemma \ref{lem:atoms_of_differences} 
an atom in the distribution of \(Y_{14}\), \(Y_{24}\) and \(Y_{34}\) each. For \(Y_{24}\), this is shown in 
Figure \ref{fig:density} where there is a peak at \(Y_{24} = 1.5\) corresponding to the value \(\omega_{24} = 1.5\).
In Figure \ref{fig:Polytrope}, we see a marginal picture of \(Y_{14}\) vs.\ \(Y_{24}\) with a horizontal boundary at 
\(Y_{14} = 3\) and a vertical boundary at \(Y_{24} = 1.5\).
\end{example}

In practice, estimating the parameter \(\omega_{ij}\) requires solving two questions, inferring the number \(K\)
of mixture components and actually estimating the parameters of the mixture.
The number of components $K$ is either known from the DAG, or may be estimated by minimizing the 
\emph{Bayesian information criterion} (BIC) \cite{Schwarz.BIC:1978}.
 
The \emph{Expectation-Maximization algorithm} (EM-algorithm) \cite{DLR:1977} is then used to infer the actual 
Gaussian mixture. 
This iterative method estimates the parameters of Gaussian mixtures consisting of \(K\) components
by iteratively refining the likelihood function. 

In the E-step of the EM-algorithm, we compute the posterior probabilities for each component given the data.
Then, in the M-step, we update the parameters \( \pi_k \), \( \mu_k \), and \( \sigma_k^2 \) 
by maximizing the expected log-likelihood. This continues until we converge.

We apply this process to a sample of \(Y_{ij}\) to obtain the mixture components as described in 
Theorem \ref{lem:distribution-of-coordinate-differences} and obtain \(\omega_{ij}\) as the minimum of the means \(\mu_k\)
according to Corollary~\ref{cor:smallest-peak}.

\begin{figure}[bt]
    \centering
    \includegraphics[width=0.7\linewidth]{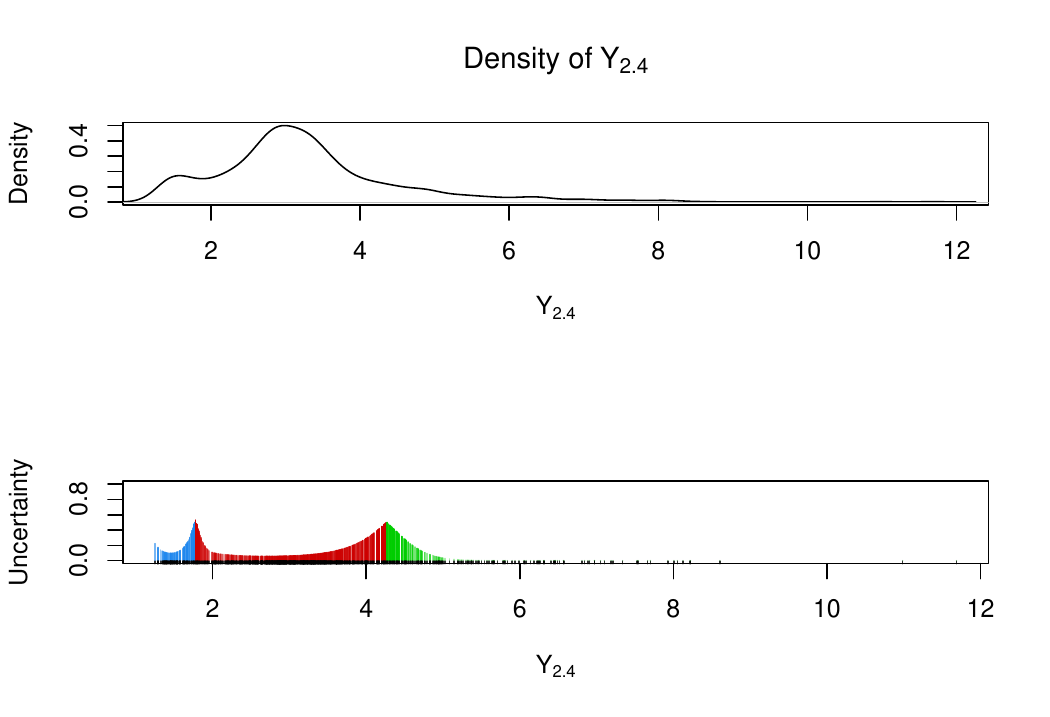}
    \caption{Density plot of \(Y_{24}\) from Example~\ref{ex:GMM} together with uncertainty of assigning
    observations to one of the $K$ components of the Gaussian mixture. 
    Assignment uncertainty is maximized at the meeting of two mixtures.}
    \label{fig:EM_Algo_atoms}
\end{figure}
    
Figure \ref{fig:EM_Algo_atoms} displays the uncertainty of atom assignment by the EM logarithm. 
Iteratively updating the probability of an observation belonging to the \(k\)-th mixture component
ensures we avoid inconsistencies caused by incomplete or data with noise.

\subsection{Geometric Estimation}\label{sec:optimization}
There is another way we can make use of the geometry of the polytrope associated to a MLBN. In particular,
this polytrope defines the support of a MLBN, meaning the support has a boundary, and by 
Corollary~\ref{cor:smallest-peak} this becomes a soft boundary.  

Since the facets of any polytrope are defined by \(Y_{ij}\)-hyperplanes, 
we can also estimate best-fit hyperplanes for the boundary of the support, turning the question of parameter estimation
in Corollary~\ref{cor:smallest-peak} into an optimization problem. 
This point-of-view is advantageous when an edge is close to structural inactivation, or when the sample size \(N\)
is small.

For a given sample $X^1, \ldots , X^N$ with $X^\nu =(X_1^\nu, \ldots , X_n^\nu)$ for $1\leq\nu\leq N$, 
we need to solve the following optimization problem for $i < j $ and $i, j \in V(G)$ and 
$\nu = 1, \ldots N$, where $\omega_{ij} \in \mathbb{R}$ and $\delta_{ij}^\nu \geq 0$ are decision variables:
\begin{optimproblem}
    \objectivefunction{$K_1\cdot \sum_{\nu=1}^N \sum_{i< j \in V(G)} \delta_{ij}^\nu + K_2\cdot \sum_{i < j \in V(G)} \omega_{ij}^2$}
    \variables{$\delta^\nu \in \mathbb{R}^{n\times n}, \nu \in [N]$ and $\omega\in\R^{n\times n}$}
    \constraints{$Y_{ij}^\nu \leq \omega_{ij} + \delta_{ij}^\nu$ and $\delta_{ij}^\nu \geq 0$}
\end{optimproblem}\noindent
This is a dual optimization problem where the linear part of the constraints are known from \eqref{eq:wdp}
and the constants need to be found.
The tuning parameters $K_1$ and $K_2$ allow us to put different emphasis on sharp boundaries in lieu of
Lemma \ref{lem:atoms_of_differences} compared to noisy, soft boundaries that are in line with Corollary \ref{cor:smallest-peak}.

\begin{example}\label{ex:hyperplanesx3}
    \begin{figure}[bht]
        \centering
        \includegraphics[trim={0 0 0 1cm},clip,width=1.0\linewidth]{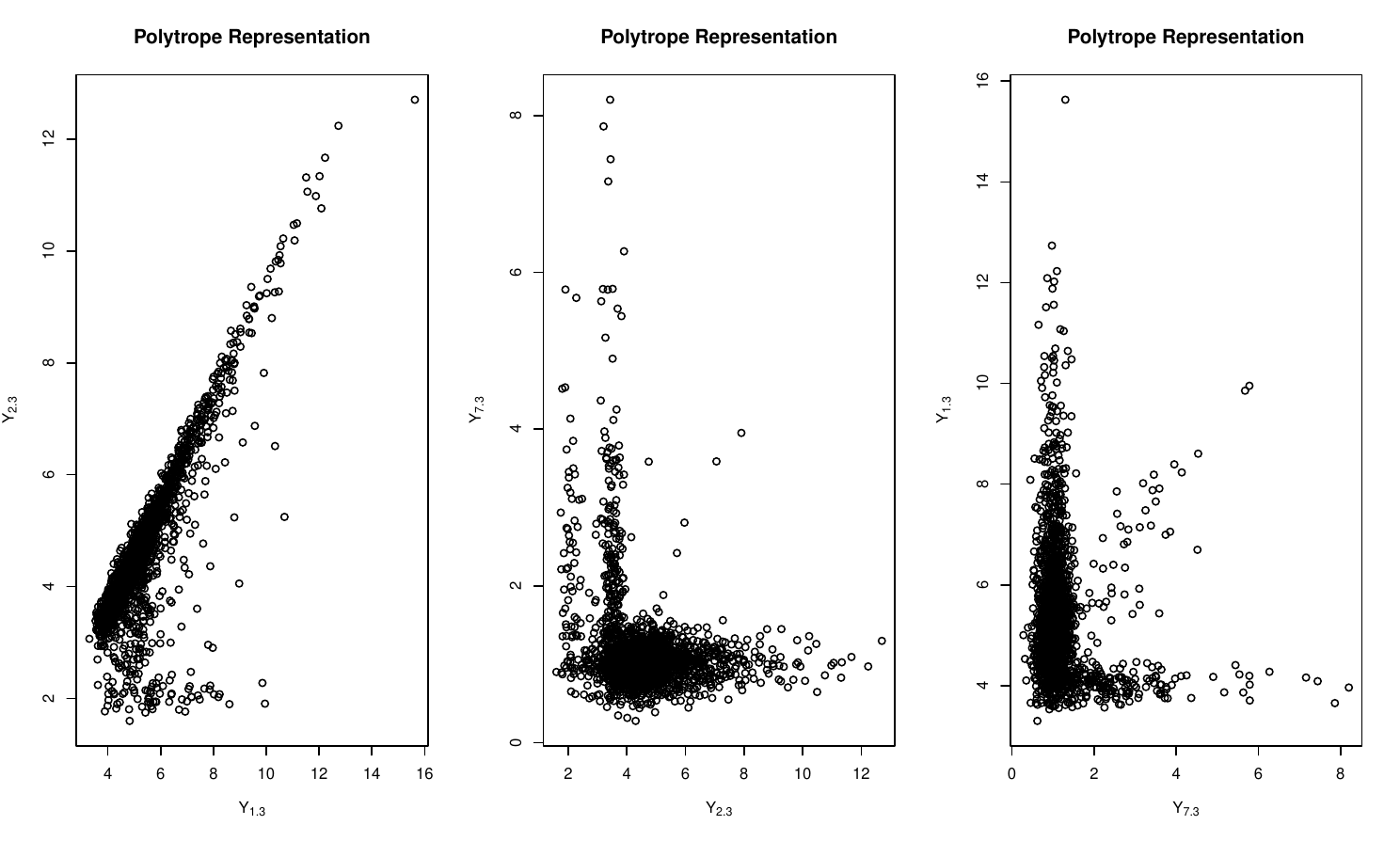}
        \caption{Selection of marginal samples from the network in Figure~\ref{fig:10node} corresponding to  polytrope facets used to infer values of $\omega_{ij}$. }
        \label{fig:HyperPlane_X3v1}
    \end{figure}
    Figure \ref{fig:HyperPlane_X3v1} shows three marginal pictures for the network from Figure \ref{fig:10node}.
    In the marginal for \(Y_{13}\) vs.\ \(Y_{23}\), we can observe a horizontal boundary at 
    \(Y_{23} = 2\) which corresponds to the estimate \(\hat\omega_{23} = 2\).
    Likewise, there is a vertical boundary at \(Y_{13} = 4\), again corresponding to the estimate \(\hat\omega_{13} = 4\).
    Additionally, there is another diagonal boundary owing to the edge \(1\to 2\).

    The boundary corresponding to \(Y_{13} = 4\) is challenging to discern in the first marginal pictures.
    However, in the marginal picture for \(Y_{73}\) vs. \(Y_{13}\), we get a much more pronounced 
    horizontal boundary, making \(\hat\omega_{13}\) easier to recover.

    In the marginal picture for \(Y_{23}\) vs.\ \(Y_{73}\), we can even observe two vertical `boundaries'
    and one horizontal boundary. These are a consequence of Theorem~\ref{lem:distribution-of-coordinate-differences} 
    where there are multiple mixture components corresponding to multiple ancestors exerting a causal influence over 
    the vertex \(3\).
\end{example}

\begin{remark}
    The upshot from Example~\ref{ex:hyperplanesx3} is that the position within the polytrope 
    from which the hyperplane $Y_{ij}$ is observed is as critical to estimation as is the choice of tuning parameters. 
    Certain regions of the polytrope may afford greater visual separation between mixture components or cleared delineation
    of hyperplane boundaries. 
    Thus, careful consideration of the observation geometry within the polytrope is essential for robust and meaningful inference. 
\end{remark}

\section{Computational Experiments}\label{sec:experiments}

In this section, we analyze components within the structure design of Figure~\ref{fig:10node}. 
This structure facilitates a comprehensive analysis of how various network configurations influence parameter recovery
while providing insight into the robustness of our estimation methods in the presence of noise. 
We provide algorithms for parameter recovery using GMM-based estimates, and hyperplane-based estimates. 
Finally, we detail how sample size and the structural activity of each edge influences the choice of estimation method.

We impose the following assumptions throughout our analysis. 
The innovations \(Z_i\) are modeled as i.\,i.\,d.\ random variables following a Fréchet distribution 
with a common location parameter $\alpha$, scale $\beta$, and shape $\xi$; 
that is, for each innovation we have \[ Z_i \sim \text{Fr\'echet}(\alpha, \beta = 1, \xi = 1),  \text{ for some constant } \alpha\in\R.\] 
Additionally, the standard deviation of the noise terms $\varepsilon_i$ as defined in Section~\ref{sec:param-estimation} 
are constrained to lie in the open interval $(0, .25]$ for all $i$, i.e., \[ 0 < \sigma_i \leq .25.\]

\subsection{Methods}
For the actual comparison, we opted to implement both approaches using 
the \texttt{R} programming language \cite{Rlang} due to the rich support of statistical and numerical workflows.
\subsubsection{Gaussian mixture models using \texttt{mclust}}
The \texttt{mclust} package by \citeauthor{mclust}~\cite{mclust}
employs an approach called \textit{Model-Based Clustering} for estimating the parameters of 
an MLBN by leveraging GMMs in an unsupervised framework. 

Given that we assume no prior knowledge beyond the directional relationships in the network, 
\texttt{mclust} provides a suitable parameter estimation approach under these constraints.
To function in this context, the algorithm determines the optimal number of mixture components using 
the BIC. This leads to a statistically driven process that adapts 
to the underlying structure without additional assumptions. 
On the other hand, \texttt{mclust} does allow for use of prior knowledge about the graph structure to improve results further. 

\subsubsection{Hyperplane fitting using \texttt{quadprog}}
The \texttt{quadprog} package by \citeauthor{quadprog}~\cite{quadprog} provides a framework for solving quadratic 
programming problems subject to linear constraints. 
This package allows us to estimate parameters in the presence of hyperplane constraints
by optimizing a quadratic objective functions while enforcing necessary geometric conditions. 
In particular, we use the function \texttt{solve.QP}. This routine implements the dual method of 
\citeauthor{Goldfarb.Idnani:1982}~\cite{Goldfarb.Idnani:1982,Goldfarb.Idnani:1983}
for solving quadratic programming problems of the form:
\begin{optimproblem}
    \objectivefunction{$-d^T b + \frac{1}{2} b^T D b$}
    \variables{$b\in\R^{N+1}$}
    \constraints{$A^T b \geq b_0$}
\end{optimproblem}

This formulation provides an efficient framework for handling our optimization problem from 
Section~\ref{sec:optimization}, as it ensures that slack variables remain strictly non-negative. 
However, in our specific formulation, the objective function \( \omega_{ij}^2 \) imposes a lower bound at zero for 
\( \omega_{ij} \), preventing the accurate estimation of the true parameter. 
We address this limitation in Algorithm~\ref{alg:hyperplane-fitting}
by shifting the observations of \(Y_{ij}\) by the minimal observation, thus making the data
being bounded by zero from below.

This adjustment ensures that the quadratic optimization functions properly and it allows the user 
to manually fine-tune the estimate based on the marginal visualization of the polytrope.
However, this method requires the user to determine the sample mean while modifying the tuning parameters. 
\begin{algorithm}[t]
\caption{Estimation of the parameter \(\omega_{ij}\) using quadratic programming}
\label{alg:hyperplane-fitting}
\begin{algorithmic}[1]
\Input Sample $\mathcal{S}:= \{X^1, \ldots , X^N\}$ originating from an MLBN
\Parameter threshold \(t>0\), tuning parameters \(K_1,K_2\geq 0\) with \(K_1 + K_2 = 1\)
\State \(Y_{ij} \leftarrow (\log{X_j^\nu} - \log{X_i^\nu})_{\nu = 1}^N\)
\State \(Y'_{ij} \leftarrow Y_{ij} - \min_\nu(Y_{ij}^\nu)\)
\State Construct the matrices \(K_2D\), \(A\) and vectors \(K_1d\) and \(b_0\)
\State Initialize \(b \coloneqq (\omega'_{ij}, \delta^1_{ij},\dots,\delta^N_{ij})\)
\While{\(\omega'_{ij} > t\)}
\State Update \(b\) using \texttt{solve.QP}
\State Adjust tuning parameters $K_1$ and $K_2$ 
\EndWhile
\State\Return \(\omega'_{ij} + \min(Y_{ij})\)
\end{algorithmic}
\end{algorithm}

\begin{figure}[b]
    \centering
    \includegraphics[width=0.9\linewidth]{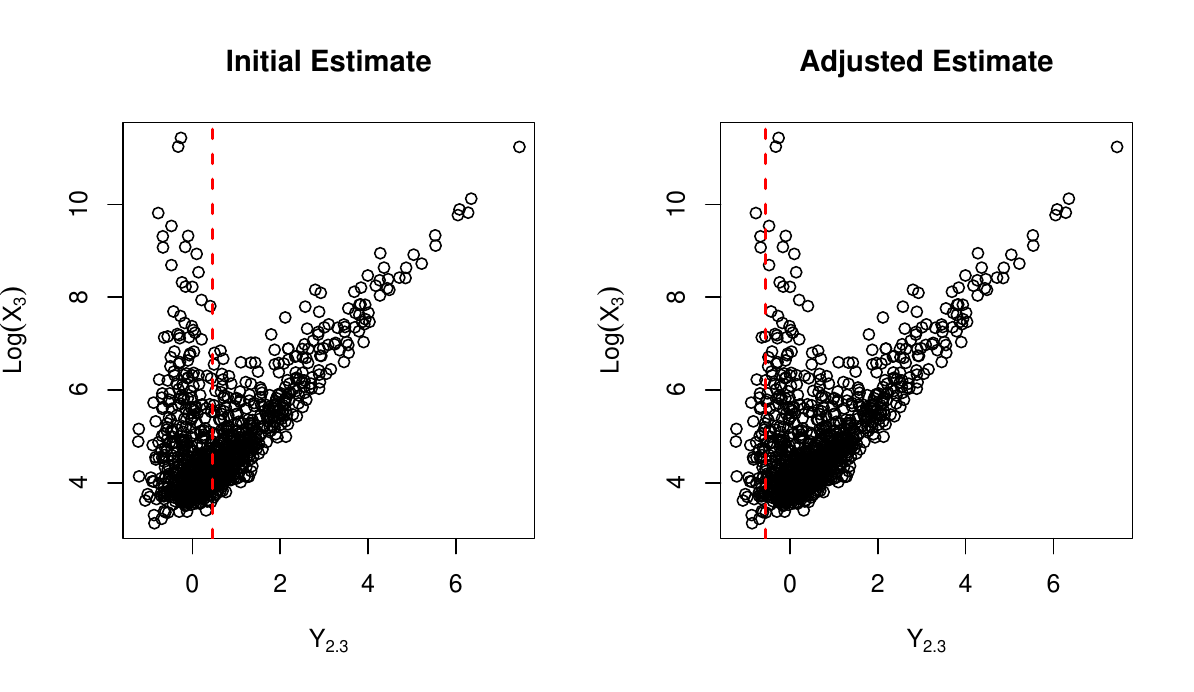}
    \caption{Following the preliminary estimation phase in Algorithm~\ref{alg:hyperplane-fitting} the facet $Y_{ij}$ is subjected to a visual inspection to inform subsequent parameter tuning. The parameters $K_1$ and $K_2$ are then manually adjusted to enhance the estimation accuracy of $\omega_{ij}$.}
    \label{fig:Hyper_estimate}
\end{figure}
    
Figure \ref{fig:Hyper_estimate} displays an initial attempt at estimating $\omega_{23} = -0.5$
in the network from Figure~\ref{fig:10node}, and the estimate following manual tuning to shift the estimate into the center of the leftmost boundary.

\subsection{Limitations of Estimation Methods}
In the formulation of the max-linear Bayesian network without noise, the estimator $\hat\omega_{ij} = \min_\nu\{Y_{ij}^\nu\}$, taken over samples where the minimum value is attained multiple times, can yield an exact parameter recovery. However, when noise, $\varepsilon_i \sim N(0,\sigma_i)$ and $\varepsilon_j \sim N(0,\sigma_j)$, are introduced to the model, the probability of observing an exact equality $\mathbb{P}_{\varepsilon_i, \varepsilon_j}(Y_{ij}+(\varepsilon_j - \varepsilon_i) = y_{ij}) = 0$ due to the continuous atom-free nature of the noise distribution. As a result, the original estimator fails to provide accurate estimates under conditions with noise, motivating the development of our alternative estimator. 

The effectiveness of parameter estimation using GMMs is contingent upon the sample size and the extent to which each mixture component $Y_{ij}$ is adequately represented. 
When the sample size is sufficiently large and $Y_{ij}$ adequately represented, GMMs yield accurate and stable parameter estimates, 
as the EM-algorithm has ample data to iteratively refine its estimates with confidence. The abundance of samples enables the model to effectively distinguish the mixture components, enhancing the reliability and robustness of results. 

However, as the sample size decreases or if $i \to j$ is not adequately represented, the performance of GMM deteriorates, and the hyperplane method emerges as a more reliable alternative.
This limitation is particularly notable in cases where a path in the network approaches structural inactivation. 
As formalized in Remark \ref{def:structural-inactivation} this scenario corresponds to the case where fewer than 5\% of samples for $X_j$ are inherited from $X_i$. In these under-represented components,the GMM framework fails to accurately estimate the corresponding edge parameter. In contrast, the hyperplane method remains effective, due to its reliance on geometric constraints, which provide more stable and reliable estimates. 

\begin{figure}[tb]
    \centering
    \includegraphics[width=0.8\linewidth]{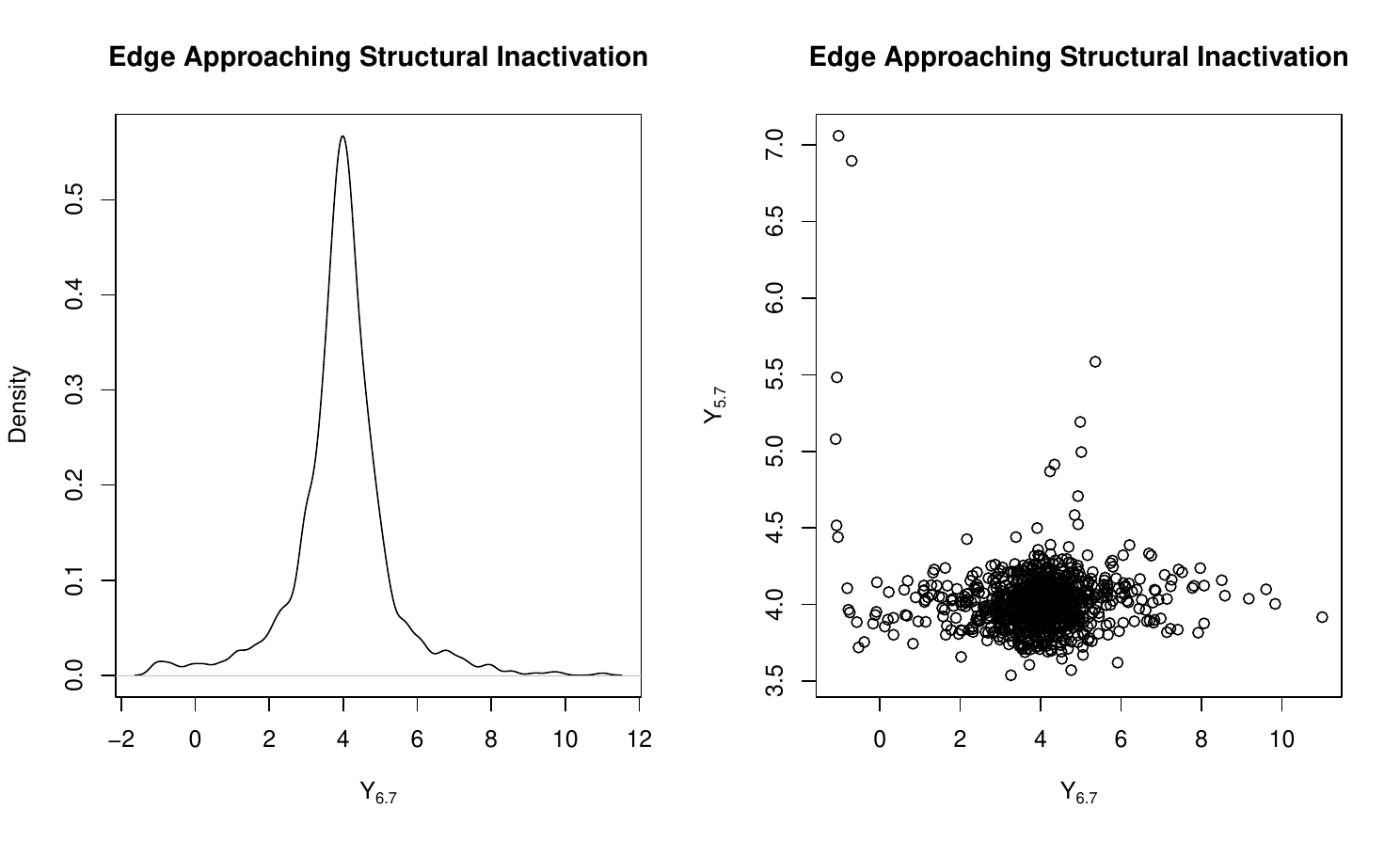}
    \caption{Structural inactivation for Subgraph III, highlighting a case where the GMM-based method fails. 
    The joint density plot of $Y_{67}$ and marginal distribution plot of $Y_{57} \text{ and } Y_{67}$ 
    are shown under the condition that the contribution of $X_{6}$ to $X_7$ is less than $ 1\%$ of $N=1000$.}
    \label{fig:Struct_Inactivation(Y)}
\end{figure}

\begin{example}
Figure~\ref{fig:Struct_Inactivation(Y)} illustrates the limiting behavior of an edge approaching structural inactivation. 
For the purposes of this analysis, we restrict attention to Subgraph III in Figure \ref{fig:GraphStructures}. 
The empirical density plot of $Y_{67}$ left exhibits heavy tailed behavior extending in both the positive and negative directions. 
This distributional characteristic is indicative of a vanishing or negligible dependency of parent vertex across an edge to its child vertex. 
Structurally, this behavior resembles that of an independent vertex, such as vertex $10$
in Figure~\ref{fig:10node}, where observational data fail to provide definitive evidence of a causal relationship. 

Under these conditions, the GMM-based procedure fails to provide a reliable estimate of the underlying components due to insufficient observations. 
Consequently, in the absence of sufficient statistical data, manual inspection of the marginal geometry becomes a pragmatic alternative. 
Specifically, in the joint distribution of $Y_{67}$ vs. $Y_{57}$ illustrated in Figure \ref{fig:Struct_Inactivation(Y)} 
a faint yet discernible vertical structure is observed, suggesting a plausible estimate  for $\omega_{67} = -1$.
\end{example}

While we have presented empirical scenarios in which the GMM fails to accurately estimate for \( \omega_{ij} \), 
we have not yet systematically discussed the conditions under which such failures occur. 
To advance our analysis of GMM limitations, we aim to quantify a critical threshold on the proportion of observations traversing edge \( i\to j \) below which the GMM-based inference becomes unreliable. 

\subsubsection{Simulation Study} 
\begin{figure}[tb]
    \centering
    \includegraphics[width=0.8\linewidth]{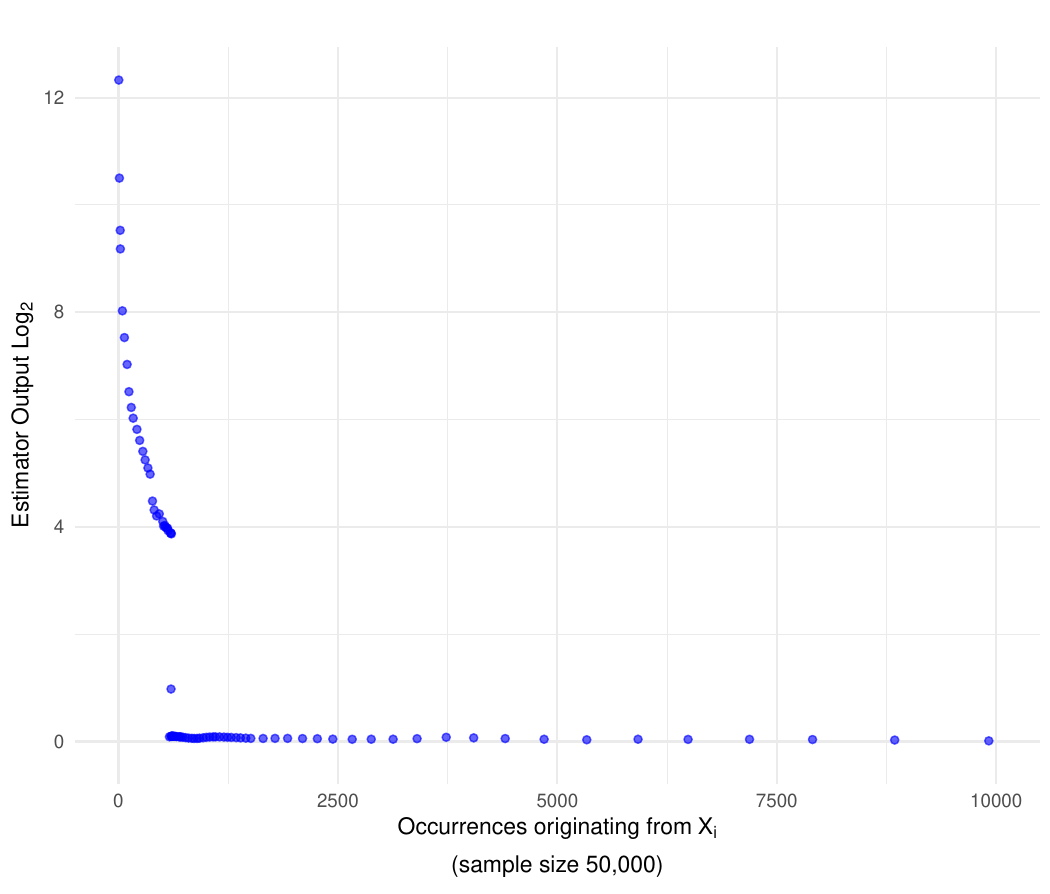}
    \caption{Global behavior of GMM estimation when $\sigma = 0.1$ and $\omega_{ij}=0$ plotted against the proportion of observations attributed to edge $i\to j$. The figure illustrates the overall stability and accuracy of the estimator, and highlights the region where tail dependency results in estimation failure.}
    \label{fig:Estimator_by_Percent+Path}
\end{figure}

In this experiment, we assume i.\,i.\,d.\ noise with standard deviation \( \sigma = 0.1 \), and a total sample size of 50,000 observations.
Figure~\ref{fig:Estimator_by_Percent+Path} reports the estimation error of 
\( \omega_{ij} \) as a function of the number of samples associated with path $i\rightsquigarrow j$. 
The vertical axis is plotted on a base-2 logarithmic scale, to facilitate interpretation of the relative magnitude of parameter estimation error when estimation fails. 

Under these conditions, the GMM-based estimator exhibits a significant upwards bias in estimating \( \omega_{ij} \) 
when fewer than 578 observations, equivalent to 1.16\% of the total sample are associated with the path \( i\rightsquigarrow j \). 
This behavior is indicative of a failure in statistical inference due to tail dependency. 
As illustrated in Figure~\ref{fig:Estimator_by_Percent+Path}, there exists a distinct threshold beyond which the GMM-based estimator ceases to be reliable. 

By fixing the noise level $\sigma$ and number $N$ of observations, we isolate a specific scenario that highlights the estimator's limitations. 
However, in a general setting, consistent estimation within the GMM framework is a function of noise levels $\sigma_i$ and the number of atoms associated with edge $i \to j$.
Elevated noise levels $\sigma$ obscure the separation between components, thereby increasing the minimum sample threshold required for accurate parameter estimation. Concurrently, the number of atoms associated with edge $i \to j$ must be sufficiently large, relative to the sample size and model complexity to guarantee reliable estimation of $\omega_{ij}$.
\begin{remark}
    The reliability of the GMM-based estimation deteriorates significantly as the noise level increases beyond $\sigma > 0.25$, due to the reduced separability of mixture components. However, the method also fails in the noiseless case, $\sigma=0$, as the resulting Dirac measures are not identified within the GMM framework. In such degenerate settings, the model violates fundamental assumptions of continuous mixture densities.
\end{remark}

\begin{figure}[bt]
    \centering
    \includegraphics[width=0.8\linewidth]{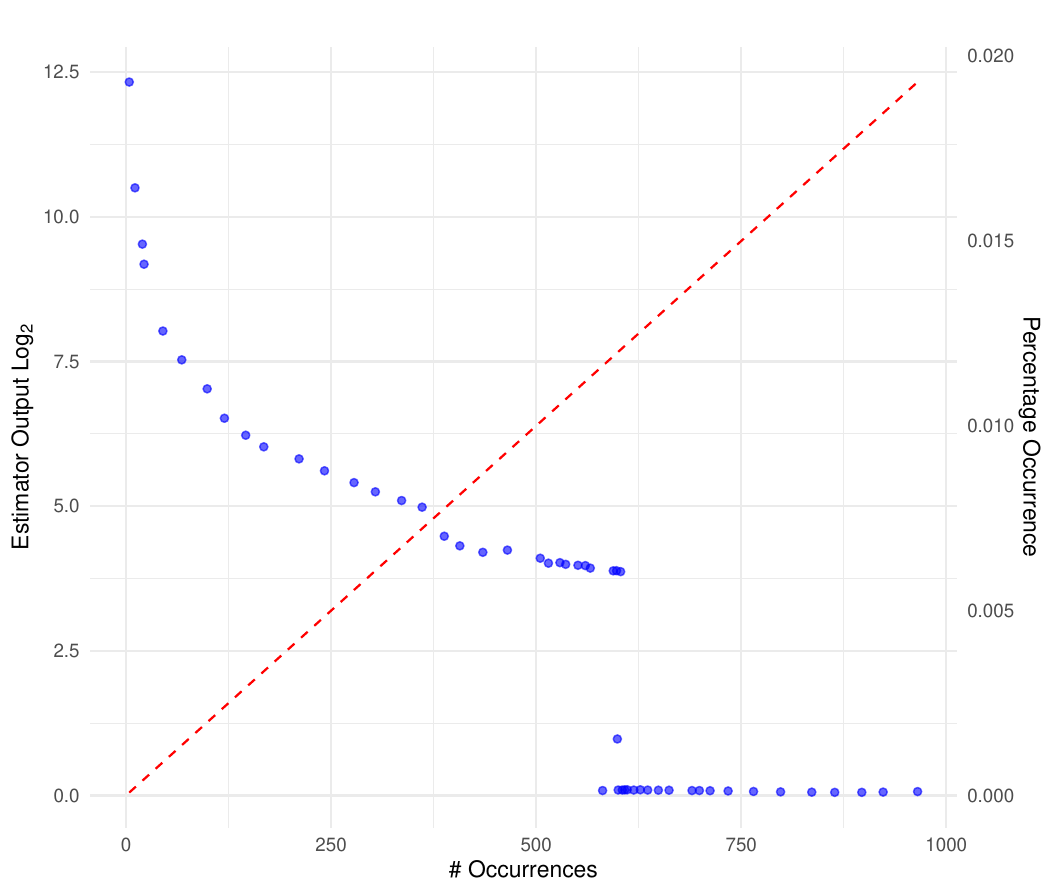}
    \caption{EM-algorithm behavior near edge inactivation: as observations diminish, probabilistic assignments of atoms to components becomes unstable leading to variance in GMM-based estimates.}
    \label{fig:Instability_of_Cij}
\end{figure}

As the GMM-based estimator approaches the boundary of its reliable parameter recovery limits, it begins to exhibit instability in both parameter estimation and assignment of mixture components. 
Figure~\ref{fig:Instability_of_Cij} provides a detailed characterization of the estimator's behavior near the critical threshold, beyond which the GMM-based method fails to consistently recover the distribution associated with $Y_{ij}$.
The red diagonal line indicates the empirical frequency (referenced on the right vertical axis), while the blue dots represent the GMM-based estimates of $\hat{\omega_{ij}}$ under the condition $\omega_{ij}=0$.
Notably, the estimator demonstrates pronounced instability when the number of atoms along edge $i\to j$ lies between 550 and 600. 
This variability is primarily attributed to the sensitivity of the EM-algorithm, which can inconsistently assign atoms to mixture components inducing non-negligible variance in the estimator.

\subsubsection{Estimator Stability} 
To further assess the reliability of the GMM-based estimation procedure, we fix the noise level at $\sigma = 0.1$ and systematically vary the sample size. The objective is to determine the minimum number of observations along edge 
$i\to j$ required for the estimator to produce stable and accurate estimates for $\omega_{ij}$. 
Table \ref{tab:observations} presents the results, listing, for each sample size, the appropriate threshold number of edge specific observations necessary for convergence.

\begin{table}[bth]
    \centering
    \begin{tabular}{ccccc}
        \toprule
        $N$ & \textbf{Path \%} & \textbf{Path obs.} & \textbf{$\hat{\omega}_{ij}$}  \\
        \midrule
        500    & 4.0\%  & 20   & 0.097  \\
        1000   & 2.3\%  & 23   & -0.002 \\
        5000   & 1.48\%  & 74   & 0.068 \\
        10000  & 1.32\% & 132  & 0.054  \\
        50000  & 1.16\% & 578  & 0.037  \\
        \bottomrule
    \end{tabular}
    \caption{Minimum number of edge-specific observations required for consistent estimation of $\omega_{ij} = 0$ under fixed noise $\sigma=0.1$. For each size $N$, the corresponding percentage of observations along edge $i \to j$, the count of such observations, and the GMM estimate are reported. Estimates are presented on a $\log_2$ scale.}
    \label{tab:observations}
\end{table}

When the GMM-based estimator fails due to an insufficient number of observations along edge $i\to j$, the tropical hyperplane method provides a more robust alternative. 
With appropriate tuning of the parameters $K_1$ and $K_2$, it becomes possible to visually assess whether the estimate aligns with the center of the vertical boundary associated with edge $i\to j$. 
In such cases, the geometry of the tropical hyperplane becomes a key diagnostic tool, especially when the edge is approaching structural inactivation. 
The spacial orientation from which the hyperplane and its corresponding facet are viewed significantly influences both the clarity and interpretability of the marginal distribution. 
This directional sensitivity highlights the value of the tropical method a theoretically grounded and practically effective approach for inference under the influence of tail dependency.

\section{Conclusion and Future Work}\label{sec:conclusion}
In this work, we added noise to a max-linear Bayesian network and presented two approaches for parameter recovery. 
Our first approach is based on a statistical framework using GMMs, while the second approach leverages the 
geometric properties of polytropes associated to MLBNs.

We give the theoretical justification for using GMMs in parameter recovery.
Additionally, we examined the performance, stability and limitations of GMM-based parameter estimation in MLBNs. 
Our analysis identified a critical threshold on the number of edge-specific observations required to ensure a statistical stable estimate.

On the other hand, we formulated a quadratic optimization problem for parameter estimation in the geometric setting.
We provide an iterative algorithm for carrying out geometric parameter estimation. 
In the end, we observed that this approach provides a more robust estimate in situations where the GMM-based 
approach failed.

\subsection{Future Directions} 
Subsequent work will aim to formalize the tropical estimation framework and extend its applicability to more general settings, including models with latent confounding, time series analysis, and hierarchical graph structures. A central theoretical goal will be to characterize estimator performance in terms of sample complexity, noise variance, and graph topology. 

\section*{Acknowledgements}
KF is funded by the Deutsche Forschungsgemeinschaft (DFG, German Research Foundation) under Germany´s Excellence Strategy 
– The Berlin Mathematics Research Center MATH+ (EXC-2046/1, project ID: 390685689).

RY and MA are partially supported by NSF Statistics Program DMS 2409819.
\printbibliography

\end{document}

%% file: figures/4node.tikz
\begin{tikzpicture}[line width=1pt, x=2cm, y=1.5cm]
    \node[draw, circle] (1) at (-1,1) {$1$};
    \node[draw, circle] (2) at (0,1) {$2$};
    \node[draw, circle] (3) at (0,0) {$3$};
    \node[draw, circle] (4) at (1,0) {$4$};
    
    \draw[->] (1) -- (2);
    \draw[->] (1) -- (3);
    \draw[->] (2) -- (3);
    \draw[->] (2) -- (4);
    \draw[->] (3) -- (4);
\end{tikzpicture}

%% file: figures/10node.tikz
\begin{tikzpicture}[scale=1,line width=1pt,x=1.25cm,y=.5cm]
    \node[draw, circle] (1) at (1,-1) {$1$};
    \node[draw, circle] (2) at (1,2) {$2$};
    \node[draw, circle] (3) at (-1,-1) {$3$};
    \node[draw, circle] (4) at (-1,2) {$4$};
    \node[draw, circle] (5) at (-2,-3) {$5$};
    \node[draw, circle] (6) at (-4,-3) {$6$};
    \node[draw, circle] (7) at (-3,-1) {$7$};
    \node[draw, circle] (8) at (-2,1) {$8$};
    \node[draw, circle] (9) at (-4,0) {$9$};
    \node[draw, circle] (10) at (-3,2) {$10$};

    \draw[->] (1) -- (2);
    \draw[->] (1) -- (3);
    \draw[->] (2) -- (3);
    \draw[->] (2) -- (4);
    \draw[->] (3) -- (4);
    \draw[->] (5) -- (7);
    \draw[->] (6) -- (7);
    \draw[->] (7) -- (3);
    \draw[->] (7) -- (8);
    \draw[->] (7) -- (9);
    \draw[->] (8) -- (4);
\end{tikzpicture}